\documentclass[letterpaper]{article} 
\usepackage{aaai24}  
\usepackage{times}  
\usepackage{helvet} 
\usepackage{courier}  
\usepackage[hyphens]{url}  
\usepackage{graphicx} 
\usepackage{natbib}  
\usepackage{caption} 
\usepackage{algorithm}
\usepackage{algorithmic}
\usepackage{algorithm}
\usepackage{algorithmic}
\usepackage{amssymb}
\usepackage{multicol}
\usepackage{multirow}
\usepackage{graphicx}
\usepackage{amsmath}
\usepackage{amssymb}

\usepackage{amssymb}
\usepackage{booktabs}
\usepackage{algorithm}
\usepackage{algorithmic}
\usepackage{multicol}
\usepackage{multirow}

\usepackage{newfloat}
\usepackage{listings}
\DeclareCaptionStyle{ruled}{labelfont=normalfont,labelsep=colon,strut=off} 
\floatstyle{ruled}
\newfloat{listing}{tb}{lst}{}
\floatname{listing}{Listing}

\title{AAAI Press Formatting Instructions \\for Authors Using \LaTeX{} --- A Guide}
\author{
    Ziqi Gu$^{1\#}$, Chunyan Xu$^{1\#}$, Zihan Lu$^{1}$, Xin Liu$^{2}$, Anbo Dai$^{2}$, Zhen Cui$^{1*}$
}
\affiliations{

    \textsuperscript{\rm 1}
    Nanjing University of Science and Technology, Nanjing, China\\
    \textsuperscript{\rm 2}
    SeetaCloud, Nanjing, China \\

}

\title{Big-model Driven Few-shot Continual Learning}

\usepackage{bibentry}

\begin{document}
\maketitle

\begin{abstract}
Few-shot continual learning (FSCL) has attracted intensive attention and achieved some advances in recent years, but now it is difficult to again make a big stride in accuracy due to the limitation of only few-shot incremental samples. Inspired by distinctive human cognition ability in life learning~\cite{science-continual,kudithipudi2022biological}, in this work, we propose a novel Big-model driven Few-shot Continual Learning (B-FSCL) framework to gradually evolve the model under the traction of the world's big-models (like human accumulative knowledge).
Specifically, we perform the big-model driven transfer learning to leverage the powerful encoding capability of these existing big-models, which can adapt the continual model to a few of newly added samples while avoiding the over-fitting problem. Considering that the big-model and the continual model may have different perceived results for the identical images, we introduce an instance-level adaptive decision mechanism to provide the high-level flexibility cognitive support adjusted to varying samples. In turn, the adaptive decision can be further adopted to optimize the parameters of the continual model, performing the adaptive distillation of big-model's knowledge information. 
Experimental results of our proposed B-FSCL on three popular datasets (including CIFAR100, minilmageNet and CUB200) completely surpass all state-of-the-art FSCL methods. 

\end{abstract}

\section{Introduction}
\footnote {$\#$~Equal Contribution.}
\footnote {*~Corresponding Author.}
Recent deep learning techniques have made remarkable strides in the field of computer vision, as evidenced by the successes achieved in some  works~\cite{KarenSimonyan2014VeryDC,he2016deep}. 
The availability of labeled data and computational power has driven advancements in machine learning. However, real-world applications often involve continuous streams of new and unseen data, posing challenges for traditional models limited to predefined label sets. 
This necessitates a neural network capable of continual learning, especially when some novel classes have few-shot samples.
Therefore, this work focuses on addressing the issues of few-shot continual learning.
They are two challenges in few-shot continual learning task:
i): a few of training samples for novel classes, which hampers effective knowledge acquisition for these classes; ii): The need to mitigate catastrophic forgetting and retain previously acquired knowledge during the continual learning process.

Recently, great attention has been paid to few-shot continual learning from various perspectives~\cite{tao2020few,chi2022metafscil,zhao2023few}.
For instance, the topology structure of the knowledge space created by different classes has been considered for few-shot continual learning with neural gas networks or graph-based models~\cite{tao2020few,zhang2021few,SonglinDong2021FewShotCL}. 
Several meta-learning approaches~\cite{chi2022metafscil,javed2019meta} have been also introduced to facilitate the preservation of previous knowledge and enable adaptation to new classes in the continual learning process.
Similarly, the self-supervised technique has  been employed to augment the model's feature extraction capability, incorporating the assistance of a self-supervised objective function in the training stage~\cite{kalla2022s3c,mazumder2021few}.
Furthermore, the forward compatible training approach~\cite{zhou2022forward} has adopted virtual prototypes to compress embeddings of learned classes and predict potential new classes, preventing forgetting these learned knowledge. 
A bilateral distillation structure~\cite{zhao2023few} has been used to alleviate the over-fitting problem when facing these few-shot labeled samples. 
In\cite{yang2023neural}, the misalignment dilemma in few-shot continual learning has been addressed by taking inspiration from the phenomenon of neural collapse.
A Gaussian kernel embedded analytic learning method~\cite{zhuang2023gkeal} has been proposed to effectively balance the preference between old and new tasks, particularly under the few-shot setting.
Although these above FCSL approaches have achieve some advantages, it is different for current network models to take a big stride with limited few-shot samples.

Inspired by the cognitive mechanism of human brain in long-life learning~\cite{science-continual,kudithipudi2022biological}, we attempt to tackle the few-shot continual learning problem from the world's big-models (such as recent fashionable (CLIP)~\cite{radford2021learning}), which can be understood as human accumulative experience. 
In this work, we propose a novel Big-model driven Few-shot Continual Learning (B-FSCL) framework, which can promote a deep neural network to progressively learn from a sequential stream of labeled few-shot samples.
Here we address the key challenges of few-shot continual learning from two perspectives.
Firstly, to harness the strong capabilities of a pre-trained big-model with a well-embedding stimulus space, we introduce a big-model driven embedding transfer module to enhance the encoding capability of the continual model.
To achieve this, we optimize the convolutional layers of the continual model by maximizing the mutual information of multi-scale feature distributions between two models. 
Thus the learned knowledge of the continual model aligns more closely with the knowledge embedded in the world's big-model. 
The big-model driven embedding transfer can facilitate the continual model to adapt to these few-shot incremental data, while alleviating the over-fitting and catastrophic forgetting problems in the FSCL process. 
Secondly, it is challenging to learn the model and obtain more accurate predictions with limited samples. 
Considering that the big-model and continual model may exhibit different inference abilities for the identical data, we introduce an instance-level adaptive decision mechanism that adjusts the inference process effectively for different samples. 
By promoting effective communication between models, the final decision can integrate the reasoning advantages of both the big-model and the continual model to achieve more accurate results.  

The primary contributions of this work are as follows:
\begin{itemize}
    \item Inspired by the cognitive mechanism of the human brain~\cite{science-continual}, we propose a novel Big-model driven Few-shot Continual Learning (B-FSCL) framework, aiming to leverage the powerful embedding and inference capabilities of the world's big-model to address the FSCL task.
    \item  We perform the progressive updating of the continual model from two aspects, namely the big-model driven embedding transfer and the instance-level adaptive decision, quickly adapting to learn new concepts especially for these few-shot incremental samples.  
\item  
Comprehensive evaluations and comparisons on three public datasets (including CUB200~\cite{CatherineWah2011TheCB}, CIFAR100~\cite{AlexKrizhevsky2009LearningML}, and miniImageNet~\cite{russakovsky2015imagenet}) well demonstrate that our B-FSCL framework 
achieves a great promising improvement when compared to existing state-of-the-art approaches in the FSCL task. 

\end{itemize}

\section{Related Work}
\subsection{Few-shot continual learning}
Few-shot continual learning has gained increasing attention in recent years, with the goal of enabling a model to learn continuously from a sequence of few-shot labeled data.
Several approaches have been proposed to address the challenges in this task, including forgetting, over-fitting, and data imbalance.
For example, Tao et al.~\cite{tao2020few} proposed to utilize a neural gas network to learn and maintain the topological structure of different category features for few-shot continual learning.
A bilateral distillation structure~\cite{zhao2023few} was proposed to alleviate catastrophic forgetting of knowledge by considering the over-fitting issue.
Inspired by the phenomenon of neural collapse, Yang et al.\cite{yang2023neural} addressed the misalignment dilemma in few-shot continual learning.
Also, Zhuang et al.~\cite{zhuang2023gkeal} adopted a Gaussian kernel embedded analytic learning method to balance the preference between old and new tasks.
Zhang et al.~\cite{zhang2021few} introduced a graph attention network based evolving classifier to facilitate information propagation between classifiers.
FACT~\cite{zhou2022forward} generates virtual classes and preserves knowledge of old classes to address the catastrophic forgetting problem. 
A bi-level optimization method~\cite{chi2022metafscil} was proposed to reduce forgetting of learned knowledge and adapt to new classes based on meta-learning.
Self-supervised stochastic classifiers were also proposed in~\cite{kalla2022s3c} to address few-shot class incremental learning using a self-supervision mechanism.
The semantic-aware knowledge distillation method~\cite{AliCheraghian2021SemanticawareKD} treats the word embedding as additional information and incorporates knowledge distillation terms to mitigate the issue of forgetting.
Michael et al.~\cite{MichaelHersche2022ConstrainedFC} proposed the constrained few-shot class-incremental learning method, which utilizes hyper-dimensional embedding to enable the continual learning of more classes beyond the fixed number of dimensions in the feature space, for the purpose of balancing the accuracy and the computational-memory cost of learning novel classes.
Further, Zhu et al.~\cite{zhu2021self} proposed an incremental prototype learning scheme to achieve continual learning by fine-tuning class prototypes using known prototypes and few-shot samples from new classes to train the model, and updating the prototypes based on this.

\subsection{Open-world model}
Inspired by the achievements of big-models of the world such as (CLIP)~\cite{radford2021learning}, SAM~\cite{kirillov2023segment}, GPT-3~\cite{dale2021gpt} and Dino~\cite{liu2023grounding}, a series of research have involved the pre-training of vision-language models on large-scale image-text datasets~\cite{radford2021learning,jia2021scaling}.
Among these approaches, the contrastive language-vision pre-training (CLIP)~\cite{radford2021learning} demonstrates impressive results across diverse downstream tasks. 
The CLIP model consists of an image encoder and a text encoder. 
During the pre-training phase, contrastive learning is employed, where a matched image-text pair serves as a positive example, while different image-text pairs are treated as negatives. 
Many studies have proposed different training methods to improve the performance of visual language models in downstream tasks. 
For instance, CLIP-FSAR~\cite{wang2023clip} introduces data augmentation methods or architectural changes to enhance the generalization performance of CLIP models on different tasks.
CLIP-Adapter~\cite{gao2021clip} is optimized for specific tasks by adding adapters at different layers in the model.
Thengane et al.~\cite{thengane2022clip} demonstrated that the zero-shot prediction of the CLIP model achieves the highest level of performance in continual learning even without any training.
ZSCL~\cite{zheng2023preventing} strengthens the correlation between images and text by improving the contrast learning strategy, leading to better performance in zero-shot transfer.
Besides, Zhuang et al.~\cite{ding2022don} adopted an adapted variant of the learning without forgetting technique which uses randomly generated sentences to calculate distillation losses.

\section{Problem Description}
Few-shot continual learning (FSCL) aims to achieve continuous learning of the model from a sequential stream of sparsely labeled samples, where data from previous tasks cannot be accessed when learning a new few-shot task. 
Formally, we define the training set, labeling set, and testing set as $X$, $Y$, and $Z$, respectively. 
A sequence of labeled training datasets $X_1$, $X_2$, $\dots$, $X_T$ is utilized for learning the model, where $X_t$ represents the training set for the $t$-th session, $Y_t$ represents the corresponding labels, and $T$ denotes the total number of continual learning sessions.
In the FSCL base session (i.e., $t=1$), the model is trained on base classes with a sufficient number of samples in the training set $X_1$ and the corresponding label set $Y_1$. 
In each continual learning session $t$ ($t>1$), the training set $X_t$ is with $N$ classes, each of which contains $K$ samples (i.e., the "N-way K-shot" configuration). 
Each training set is constructed to avoid the repetition of class labels. 
In other words, for any pair of sessions $i$ and $j$ where $i \neq j$, the label sets $Y_i$ and $Y_j$ have no overlapping classes, i.e., $Y_i \cap Y_j = \varnothing$.
For $t > 1$, the training set $X_t$ consists of few-shot samples for the novel classes introduced in the $t$-th session, and its size is smaller compared to the size of $X_1$.
The test set $Z$ is used to evaluate the classification accuracy at each session $t$, and it may contain classes from all the training label sets combined, denoting as $\{Y_1 \cup Y_2 \dots \cup Y_T\}$.

\section{The B-FSCL Method}
\subsection{Motivation}
Few-shot continual learning focuses on optimizing the model to efficiently learn and adapt to new knowledge with limited samples while retaining learned knowledge. 
There are two severe challenges we need to address: i) Scarce training data for novel classes, which hampers effective knowledge acquisition for these classes; ii) The necessity to alleviate catastrophic forgetting and preserve previously learned knowledge while undergoing the continual learning. 
In fact, humans can quickly learn each new task without forgetting previous ones, while artificial neural networks do not have such extraordinary abilities. 
After investigating the cognitive mechanisms of human brain, the studies~\cite{science-continual,kudithipudi2022biological} find that augmenting large-scale neural networks with abundant supervised/unsupervised experience could learn a good embedding of the stimulus space (similar to that observed in humans), and further reduce the catastrophic forgetting in the continual learning process. 

Inspired by this above findings, we attempt to leverage the powerful understanding capability of existing big-models of the world (such as CLIP~\cite{clip-radford2021learning}), which can be understood as the empirical knowledge that humans have learned to a certain extent. 
Here we address the FSCL task from the following two aspects.
Firstly, we can perform the big-model driven transfer learning to employ the powerful encoding capability of the pre-trained big-model, which can adapt the continual model to a few newly added samples while alleviating the over-fitting and catastrophic forgetting problems.
Secondly, considering that the big-model and the continual model may have different perception results even for the same thing, we introduce an instance-level adaptive decision to provide high-level inference support adjusted to varying image samples. 
Similar to the effective communication between people, the final decisions can well consider the inference strengths of both the big-model and the continual model under the guide of the instance-level adaptive mechanism. 
Furthermore, the results can be used to gradually optimize the parameters of the continual model, performing the knowledge distillation from the big-model of the world in the continual learning process.

\begin{figure} 
	\centering
	\includegraphics[width=1\linewidth]{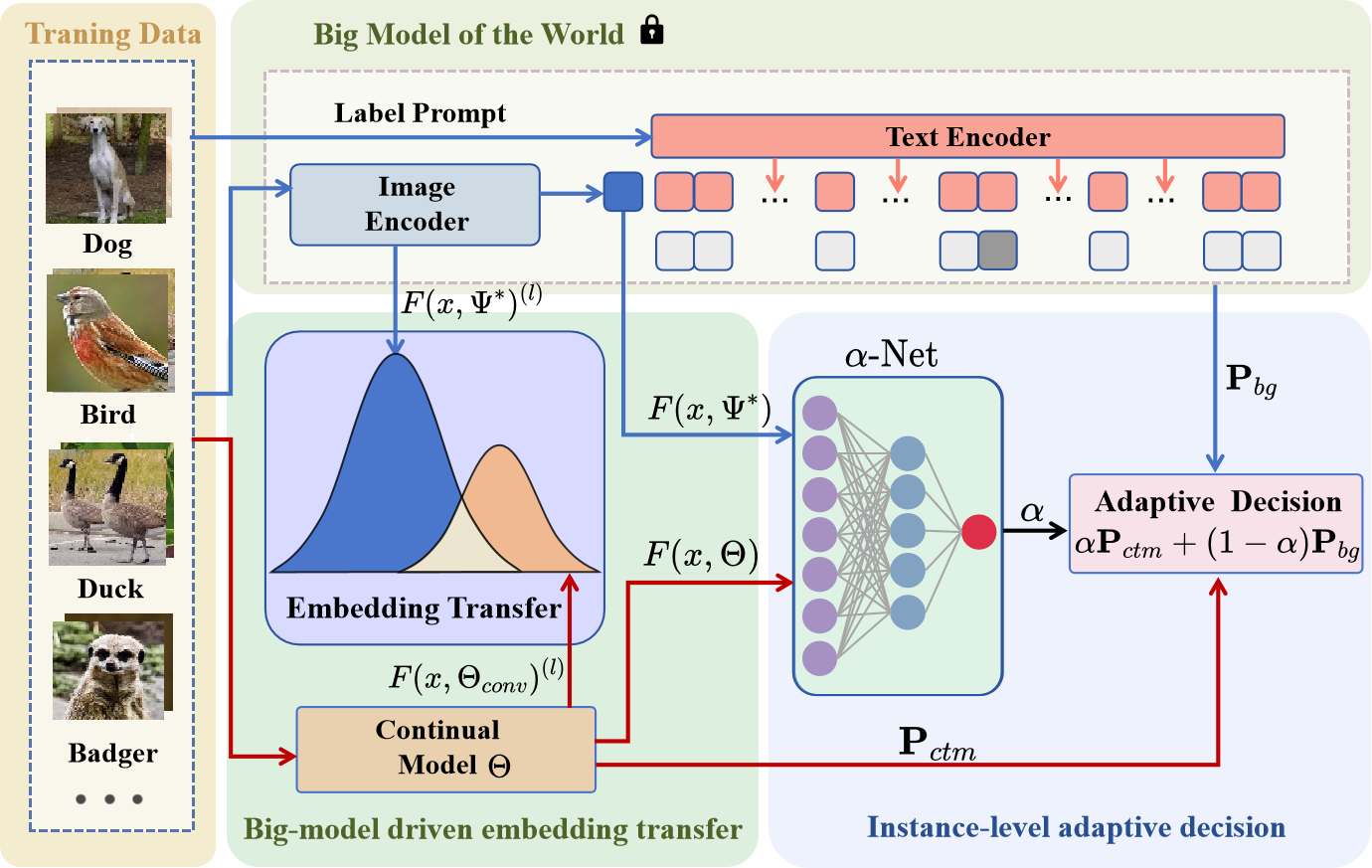} \vspace{-0.5cm}
	\caption{The pipeline of our B-FSCL framework. Under the traction of the world's big-model, we gradually optimize the parameters of the continual model (i.e., $\Theta$) in the training process. We use $\Psi^*$ to represent the big-model, because its parameters are always frozen. Here we design a big-model driven embedding transfer module and an instance-wise adaptive decision module to improve its encoding and inference capabilities when addressing the FSCL task.}
	\vspace{-0.5cm}
	\label{fig:framework1}
\end{figure}

\subsection{B-FSCL Paradigm}
Fig.\ref{fig:framework1} illustrates the pipeline of our proposed B-FSCL framework. 
Different from existing FSCL methods, the entire B-FSCL is continuously learning under the guidance/traction of the world's existing big-models, which have the powerful embedding and reasoning capabilities for the data to be processed.  
First, by taking full advantage of the world's big-model (denoting as $\Psi^*$), we design a big-model driven embedding transfer module to enhance the encoding ability of the continual model (denoting as $\Theta$), especially for the few-shot learning sessions. 
Specifically, we maximize the mutual information of two distributions between the convolutional feature  $F(x,\Theta_{conv})^{(l)}$ of the continual model and the corresponding feature $F(x,\Psi^{\ast})^{(l)}$ of big-model on multiple scales (i.e., $l=1,2, \dots, L$). 
Through the big-model driven transfer learning, we can quickly learn new concepts by adapting the continual model to new few-shot samples, while alleviating the over-fitting and catastrophic forgetting problems of the FSCL task. 
Furthermore, considering potential disparities in understanding capability between the big model and the continual model for the identical sample, we introduce the instance-level adaptive decision module to give a more reasonable prediction for different input images. 
Here we construct an $\alpha$-Net to learn an adaptive decision parameter $\alpha$ with $F(x,\Psi^{\ast})$ from the big-model and $F(x,\Theta_{conv})$ from the continual model as the input. 
Under the guidance of the learned parameter $\alpha$, we can obtain the final classification result by considering the decision probabilities of the big-model and the continual model (i.e., $\mathbf{p}_{bg}$ and $\mathbf{p}_{cmt}$). 
We use the same training strategy described above for both base and new sessions in our proposed B-FSCL framework.

\subsection{Big-model driven Embedding Transfer}
The limited learning capacity of the continual model often hampers its ability to grasp concepts from a few newly added samples.
In most existing FSCL methods\cite{zhang2021few,zhou2022forward}, a common strategy to mitigate forgetting is to only update the last fully-connected layer while freezing the encoding convolutional layers. 
However, these approaches make it difficult to learn new concepts from few-shot samples.
Inspired by the cognitive mechanism of the human brain \cite{science-continual}, we attempt to leverage the strong capabilities of a pre-trained big model that possesses a well-embedded stimulus space. 
Here we perform the big-model driven embedding transfer to enhance the encoding capability of the continual model, thereby alleviating the issues of catastrophic forgetting and over-fitting to a certain extent.

Under the traction of the world's big-model, we hope that the learned knowledge of the continual model can be more inclined to the big-model.  
Specifically, we maximize mutual information of multi-scale feature distributions as follows: 
\begin{equation}
\begin{aligned}
   \mathop{\arg\max}\limits_{\Theta_{conv}} 
    \sum_{l=1}^{L} \mathcal{I}_{\Theta_{conv}} 
    (F(x,\Theta_{conv})^{(l)}, F(x,\Psi^{\ast})^{(l)} ), 
\end{aligned}
\label{transfor}
\end{equation}
where $\Theta_{conv}$ denotes the convolutional parameters of the continual model,  $\Psi^{\ast}$ refers to the fixed big-model, and $L$ represents the total number of convolutional scales involved. 
$F(x,\Theta_{conv})^{(l)}$ denotes the convolutional feature of the input image $x$ in the $l$-th layer of the continual model, while $F(x,\Psi^{\ast})^{(l)}$ represents the corresponding convolutional feature of the big-model. 
Since the mutual information of discrete features is difficult to calculate, we employ the lower-bound to the mutual information based on the Kullback-Leibler divergence~\cite{donsker1975variational}. 
The big-model driven embedding transfer process in Eqn. (\ref{transfor}) can be further represented as: 
\begin{equation}
\begin{aligned}
   \mathop{\arg\max}\limits_{\Theta_{conv}, \varphi}  \sum_{l=1}^{L} \mathbb{E}_{\mathbb{J}}[T_{\varphi}(F(x,\Theta_{conv})^{(l)}, F(x,\Psi^{\ast})^{(l)} )] \\
    -\log \mathbb{E}_{\mathbb{M}}[e^{T_{\varphi}(F(x,\Theta_{conv})^{(l)}, F(x,\Psi^{\ast})^{(l)} )}],
\end{aligned}
\label{transfor_mini1}
\end{equation}
where $T_{\varphi}(\cdot)$ is a discriminant function between features of different models. 
$\mathbb{E}_{\mathbb{J}}$ represents the expectation of the joint distribution of $F(x,\Theta_{conv})^{(l)}$ and $F(x,\Psi^{\ast})^{(l)}$, while $\mathbb{E}_{\mathbb{M}}$ denotes the expectation of the margin distributions of $F(x,\Theta_{conv})^{(l)}$ and $F(x,\Psi^{\ast})^{(l)}$. 
Through gradually optimizing the network parameters $\Theta_{conv}$ and $\varphi$, the features of the continual model $F(x,\Theta_{conv})$ would stably converge towards the embedding  of the world's big-model $F(x,\Psi^{\ast})^{(l)}$, which also make the continual model adapt to these incremental image samples.

\subsection{Instance-level Adaptive Decision}
In the continual learning process, another major challenge is that it is difficult to learn the model and obtain accurate prediction results with only few-shot samples. 
Considering that big-model and continual model may possess varying inference abilities for the same sample, we introduce an instance-level adaptive decision mechanism to provide the effective inference adjusted to different samples. 
By promoting effective communication between models, the final decision can comprehensively integrate the reasoning advantages of both the big-model and the continual model, thereby enabling more accurate decision-making. 
To guide the instance-wise adaptive decision, we build an $\alpha$-Net to learn the weight coefficient $\alpha$ for each input, formally, 
\begin{equation}
	\alpha = \mathrm{H}(
    F(x,\Theta), F(x, \Psi^{\ast}), \omega_{\alpha}
    ),
\end{equation}
where $F(x,\Theta)$ and $F(x, \Psi^{\ast})$ represent the embedding features of the image $x$ from the continual model and the big-model, respectively. 
$\omega_{\alpha}$ denotes the parameters of $\alpha$-Net to be learned. 
The weight  $\alpha\in [0,1]$ is achieved  by considering the feature embeddings of two models (i.e., $F(x,\Theta)$ and $F(x, \Psi^{\ast})$). 
Under the guidance of the learned weight  $\alpha$, the instance-level adaptive decision can be performed as follows:
\begin{equation}
	\mathbf{p} = \alpha*\mathbf{p}_{ctm}  + (1 - \alpha)*\mathbf{p}_{bg},
\end{equation}
where $\mathbf{p}_{ctm}$ and $\mathbf{p}_{bg}$ represent the probability distributions achieved from the continual model and the big-model, respectively. 
$\mathbf{p}$ refers to the probability distribution of the final decision.

To optimize the network parameters of the continual model  and  $\alpha$-Net (i.e., $\Theta$ and $\omega_{\alpha}$), we minimize the cross-entropy value between the ground truth and the probability of our final decision: \begin{equation}
	\mathop{\arg\min}_{\Theta, \omega_{\alpha}} -\sum_i^{|X_t|}y_i \log p_i,
\end{equation}
where $i$ is the index of the training set $X_t$. 
For each input image $x_i$, $y_i$ is its ground truth and the final decision probability $p_i$  is the maximum value in the probability distribution of $\textbf{p}_i$. 
The instance-level adaptive decision allows the comprehensive integration of inference advantages between the big-model and the continual model in the ultimate decisions, achieving more effective classification results. 
Meanwhile, this process can selectively distillate knowledge from the world's big-model to the continual model, which can make the continual model adapt to these new-added samples.

\begin{table*}[!t] 
	\footnotesize
	\centering
	\caption{Performance comparison between our proposed B-FSCL and other state-of-the-art methods on the CUB200 dataset. Results marked with * are obtained from the authors' published code. } \vspace{-0.15cm}
	\scalebox{0.84}{
		\begin{tabular}{lccccccccccccccc}
			\toprule  
			\multirow{2}{*}{\bfseries Methods\mdseries} & \multicolumn{11}{c}{Accuracy in each session (\%) $\uparrow$}& \multirow{2}{*}{KR$\uparrow$}&\multirow{2}{*}{$\Delta$Final$\uparrow$}&\multirow{2}{*}{Avg$\uparrow$}\\
			\cmidrule(l){2-12} &
			
			1&2&3&4&5&6&7&8&9&10&11\\
			\midrule 
			Ft-CNN&68.68& 43.70& 25.05& 17.72& 18.08& 16.95& 15.10& 10.60& 8.93& 8.93& 8.47 & 12.33& +60.87&22.02\\
			NCM~\cite{hou2019learning} & 68.68 & 57.12& 44.21& 28.78& 26.71& 25.66& 24.62& 21.52& 20.12& 20.06& 19.87& 28.93 & +49.47& 32.49\\
			iCaRL~\cite{rebuffi2017icarl} & 68.68& 52.65& 48.61& 44.16& 36.62& 29.52& 27.83& 26.26& 24.01& 23.89& 21.16 & 30.80 &+48.18&36.67\\
			EEIL~\cite{castro2018end}& 68.68& 53.63& 47.91& 44.20& 36.30& 27.46& 25.93& 24.70& 23.95& 24.13& 22.11& 32.19 &+47.23& 36.27\\
			TOPIC~\cite{XiaoyuTao2020FewShotCL}& 68.68& 62.49& 54.81& 49.99& 45.25& 41.40& 38.35& 35.36& 32.22& 28.31& 26.28& 38.26 &+43.06 & 43.92 \\
			SPPR~\cite{zhu2021self}&68.68& 61.85& 57.43& 52.68& 50.19& 46.88& 44.65& 43.07& 40.17& 39.63& 37.33 & 54.35 &+32.01& 49.32\\
			D-DeepEMD~\cite{zhang2020deepemd}&  75.35& 70.69& 66.68& 62.34& 59.76& 56.54& 54.61& 52.52& 50.73& 49.20& 47.60 & 63.17 &+21.74& 58.73\\
			D-NegCosine~\cite{liu2020negative}& 74.96& 70.57& 66.62& 61.32& 60.09& 56.06& 55.03& 52.78& 51.50& 50.08& 48.47 & 64.66 &+20.87 & 58.86\\ 
			D-Cosine~\cite{vinyals2016matching}& 75.52& 70.95& 66.46& 61.20& 60.86& 56.88& 55.40& 53.49& 51.94& 50.93& 49.31 & 65.29 &+20.03 & 59.36\\
			CEC~\cite{zhang2021few}&75.80&71.94&68.50&63.50&62.43&58.27&57.73&55.81&54.83&53.52&52.28 & 68.97 &+17.06 & 61.33\\ 
			Mate-FSCIL~\cite{chi2022metafscil}&75.90&72.41&68.78&64.78&62.96&59.99&58.30&56.85&54.78&53.83&52.64 & 69.35 &+16.70 & 61.93\\ 
			FACT*~\cite{zhou2022forward} &77.38&73.91&70.32&65.91&65.02&61.82&61.29&59.53&57.92&57.63&56.46 & 72.95&+13.68 & 64.29\\ 
            GKEAL\cite{zhuang2023gkeal}&	78.88&	75.62&	72.32&	68.62&	67.23&	64.26&	62.98&	61.89&	60.20&	59.21&	58.67&74.38&+10.67&66.35\\
            NC-FSCIL\cite{yang2023neural}&	80.45&	75.98&	72.30&	70.28&	68.17&	65.16&	64.43&	63.25&	60.66&	60.01&	59.44&73.88&+9.90&67.28\\
            ALICE\cite{peng2022few}&	77.40&	72.70&	70.60&	67.20&	65.90&	63.40&	62.90&	61.90&	60.50&	60.60&	60.10&77.65&+9.24&65.75\\
            CABD\cite{zhao2023few}&	79.12&	75.63&   73.21&	69.93&	68.32&	66.30&	65.15&	64.96&	64.20&	64.03&	63.81&80.65&+5.53&68.61\\
			\midrule
            B-FSCL(Ours)&\textbf{81.64}& \textbf{79.45}& \textbf{77.29}& \textbf{72.85}& \textbf{73.54}& \textbf{71.86}& \textbf{71.83}& \textbf{70.16}& \textbf{69.55}& \textbf{68.93}& \textbf{69.34}&\textbf{84.93}&-&\textbf{73.31}\\
                
			\midrule  
            \emph{Improvement over CABD} &+2.52& +3.82& +4.08& +2.92& +5.22& +5.19& +5.56& +6.68& +5.20& +4.90& +5.53&+4.28&-&+4.70 \\
			\bottomrule            
		\end{tabular}
		\label{tb:CUB200}  
	} \vspace{-0.3cm}
\end{table*}

\section{Experiment}

\subsection{Experimental Setup}

\textbf{Datasets:}
We evaluate the performance of B-FSCL on three publicly available datasets: CIFAR100~\cite{AlexKrizhevsky2009LearningML}, miniImageNet~\cite{russakovsky2015imagenet}, and CUB200~\cite{CatherineWah2011TheCB}.
CIFAR100~\cite{AlexKrizhevsky2009LearningML} consists of 100 classes with a total of 60,000 RGB images.
miniImageNet~\cite{russakovsky2015imagenet} is a subset of ImageNet, comprising 100 classes. 
CUB200~\cite{CatherineWah2011TheCB} is a fine-grained classification dataset with 11,788 images distributed among 200 classes.
For the CIFAR100 and miniImageNet datasets, we use all training data of the 60 base classes to train a base network. 
The 40 new classes are then used for eight 5-way 5-shot continual learning tasks.
We divide the 200 classes of CUB200 into 100 base classes and 100 new classes. 
The 100 new classes are used for ten 10-way 5-shot continual learning tasks. 
To ensure a fair comparison, we follow the same split setting on the three datasets, as in FSCIL~\cite{XiaoyuTao2020FewShotCL}. 

\begin{figure} \hspace{-0.3cm}
	\centering
	\footnotesize
	\includegraphics[width=1.03\linewidth]{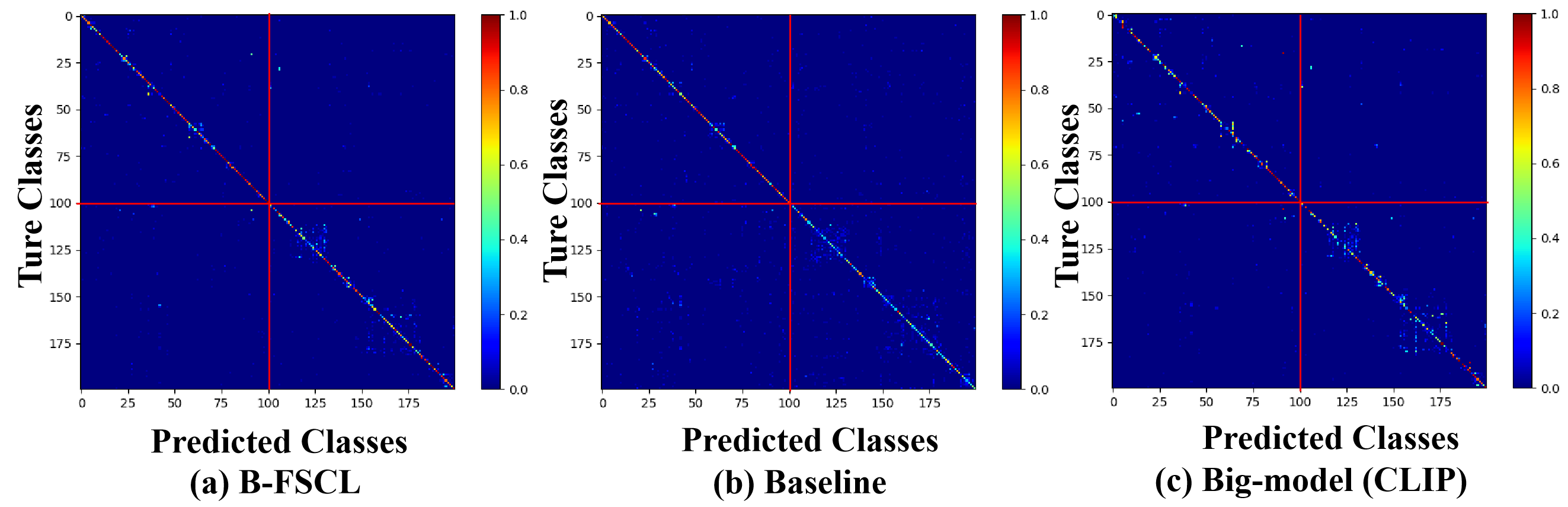} \vspace{-0.5cm} 
	\caption{Confusion matrix of the final results on the CUB200 dataset. (a) B-FSCL. (b) Baseline. (c) the Big-model (CLIP). We delineate the boundary between the base class and the new class with red lines. Our approach improves the model's predictive performance, leading to a more concentrated.}
	\vspace{-0.5cm}
	\label{fig:matrix}	
\end{figure}

\noindent\textbf{Implementation Details:}
Referring to FSCIL~\cite{XiaoyuTao2020FewShotCL}, we employ ResNet20 as the backbone for CIFAR100~\cite{AlexKrizhevsky2009LearningML}, while ResNet18 serves as the backbone for both miniImageNet~\cite{russakovsky2015imagenet} and CUB200~\cite{CatherineWah2011TheCB}.
The mutual information estimate network $\varphi$ consists of three convolutional layers with kernel sizes of 1, 3, and 5.
The $\alpha$-network consists of three fully connected layers with a sigmoid activation function and outputs the measurement of $\alpha$ for each sample.
In our experiments, we select CLIP~\cite{clip-radford2021learning} as the world's big-model that adopts 'ViT-L/14' as the pre-trained model and the "The image depicts a \{\}" as prompt.
In the training stage, we incorporate data augmentation techniques, including random cropping, random scaling, and random horizontal flipping.
To evaluate the performance of our proposed method comprehensively, we employ multiple metrics.
"$\text{Acc}_t$" represents the Top-1 accuracy in the $t$-th continual stage, while "Avg" denotes the average score, calculated as $\sum_{t=1}^T\text{Acc}\_t / T$.
"$\Delta$Final" indicates the difference in $\text{Acc}\_t$ between our method and the compared method in the final stage.
"KR" stands for the knowledge retention rate, defined as $\text{Acc}\_T/\text{Acc}\_1$.
All experiments are conducted by the PyTorch framework on NVIDIA GeForce 4090 GPU.
Due to space limitation, some experimental results are deferred to the supplementary material.
\begin{table*}[!t]
	\footnotesize
	\centering
	\caption{Comparisons of our method using different components on the CUB200 dataset.} 
  \vspace{-0.2cm}
	\scalebox{0.84}{
		\begin{tabular}{cccccccccccccccccc}
			\toprule  
			\multirow{2}{*}{ Baseline} &\multirow{2}{*}{Big-model (CLIP)} &\multirow{2}{*}{BET} &\multirow{2}{*}{IAD}& \multicolumn{11}{c}{Accuracy in each session (\%) $\uparrow$}& \multirow{2}{*}{KR}&\multirow{2}{*}{Avg}\\
			\cmidrule(l){5-15}
			&&&&1&2&3&4&5&6&7&8&9&10&11\\
			\midrule  
			&\checkmark && & 64.27&	60.44&	59.42&	56.84&	58.71&	58.41&	57.89&	54.99&	54.60&	55.58&	57.02 & - & - \\ 
			\midrule  
			\checkmark& && & 79.42&	75.65&	71.35&	66.88&	66.13&	63.99&	62.84&	60.73&	60.23&	59.45&	58.32& 73.43&65.91 \\ 
			\checkmark& &\checkmark& &79.92&	77.24&	74.11&	69.57&	69.62&	67.16&	66.12&	64.75&	63.45&	62.57&	61.67 & 77.65&68.74 \\
			\checkmark&&  &\checkmark& 81.03&	77.99&	75.52&  70.69& 	71.89&	 70.47&	69.64&   68.07&	66.11&	66.08&	66.26 & 81.77&	71.25\\
			\checkmark& &\checkmark & \checkmark &81.64& 79.45& 77.29& 72.85& 73.54& 71.86& 71.83& 70.16& 69.55& 68.93& 69.34 &84.93 &73.31\\
			\bottomrule 
		\end{tabular} \vspace{-0.8cm}
		\label{tb:ablation}
	}
\end{table*}

\begin{figure*}
	\footnotesize
	\centering
	\includegraphics[width=1\linewidth]{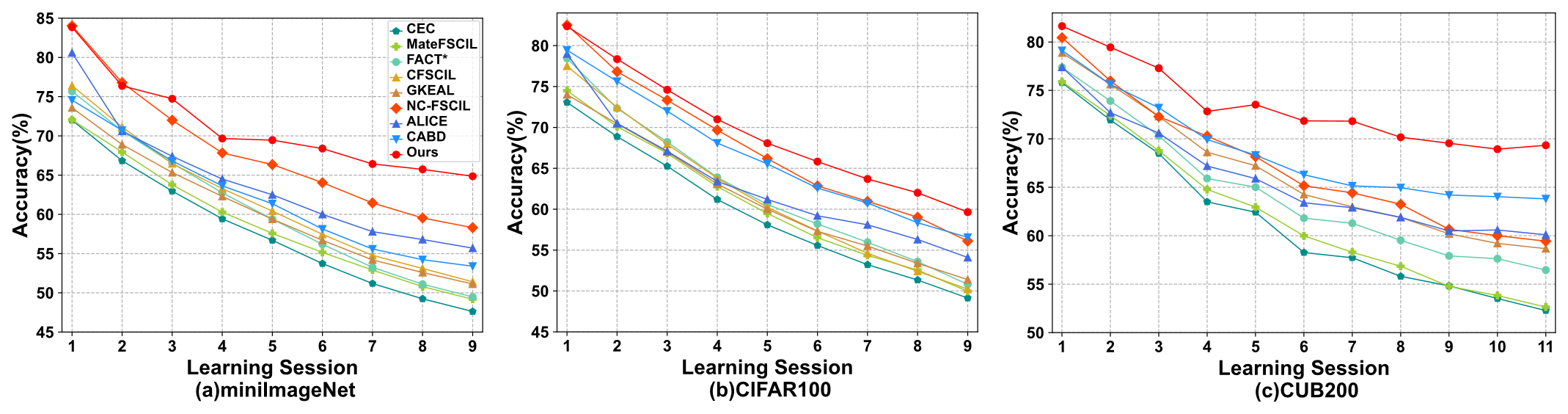} \vspace{-0.5cm} 
	\caption{
The accuracy changing curves depicting the continual sessions of different methods on the three datasets are shown in the three images. 
These images share a common legend, which is presented in the CUB200 dataset ~\cite{CatherineWah2011TheCB}.
	}
	\vspace{-0.5cm}
	\label{fig:result}
\end{figure*}

\subsection{Comparison with state-of-the-art methods}
To assess the superiority of our proposed B-FSCL framework, we conduct an evaluation on the CUB200 dataset~\cite{CatherineWah2011TheCB} and compare its performance with other state-of-the-art methods. 
The detailed results are presented in Table~\ref{tb:CUB200}, and the accuracy curves across continual learning sessions are illustrated in Fig.~\ref{fig:result}(c).
Our B-FSCL framework shows an impressive performance, significantly outperforming all other compared methods. 
Notably, our B-FSCL demonstrates significant improvements in each stage compared to the second-best approach CABD~\cite{zhao2023few}, which validates the superiority of our method.
In the final continual stage, our approach significantly outperforms CABD by a $\Delta$Final of 5.53\%.
This result highlights the consistent learning capability of B-FSCL throughout the continual stage.
We confirm that the big-model driven embedding transfer module can effectively transfer the powerful encoding ability of big-model to the continual model, thereby addressing the issue of insufficient learning capability for new classes and improving adaptability to new tasks with few-shot samples.
Compared to the current best method (i.e., CABD\cite{zhao2023few}), our proposed B-FSCL achieves 4.70\% in terms of the Avg. metric.
It indicates that our B-FSCL has superior performance across all continual learning stages.
This finding further confirms that the proposed B-FSCL can effectively mitigate the lack of the learning capacity of the continual model. 
Impressive results in Avg and $\Delta$Final metrics further verifies the superiority of our B-FSCL in learning new knowledge and alleviating catastrophic forgetting. 

Our experimental results consistently demonstrate that B-FSCL outperforms all other methods, surpassing the best CABD\cite{zhao2023few} method with 4.28\% in the KR metric.
The KR metric measures the knowledge ratio between the final continual stage and the base stage, where a higher KR value indicates better retention of learned knowledge in continual learning stages.
Moreover, these results that the B-FSCL can effectively enhance the few-shot classification accuracy by integrating the big-model driven embedding transfer and instance-level adaptive decision modules.
Our proposed B-FSCL demonstrates outstanding performance by integrating the reasoning advantages of both the big-model and the continual model, resulting in more accurate decision-making.
Additionally, the accuracy changing curves depicted in Fig.~\ref{fig:result}(c) visually validate the superiority of our B-FSCL framework.
Besides, we incorporate distinct accuracy curves for the miniImageNet and CIFAR100 datasets in Fig.~\ref{fig:result}(a) and Fig.~\ref{fig:result}(b), respectively.
The evaluation results clearly demonstrate the superior performance of our B-FSCL method across different datasets.
This further emphasizes the effectiveness of our proposed big-model driven embedding transfer and instance-level adaptive decision modules in acquiring knowledge of novel classes while addressing the issue of catastrophic forgetting.

\begin{table*}[!t]
	\footnotesize
	\centering
	\caption{Performance comparison of different methods with the similar base performance.} \vspace{-0.15cm}
	\scalebox{0.84}{
		\begin{tabular}{lcccccccccccccc}
			\toprule  
			\multirow{2}{*}{\bfseries Methods\mdseries} & \multicolumn{11}{c}{Accuracy in each session (\%) $\uparrow$}& \multirow{2}{*}{KR$\uparrow$}&\multirow{2}{*}{$\Delta$Final$\uparrow$}&\multirow{2}{*}{Avg$\uparrow$}\\
			\cmidrule(l){2-12} &
			1&2&3&4&5&6&7&8&9&10&11\\
			\midrule 
			ALICE\cite{peng2022few}&	77.40&	72.70&	70.60&	67.20&	65.90&	63.40&	62.90&	61.90&	60.50&	60.60&	60.10&77.65&+3.43&65.75\\
			FACT*~\cite{zhou2022forward} &77.38&73.91&70.32&65.91&65.02&61.82&61.29&59.53&57.92&57.63&56.46 & 72.95&+7.07 & 64.29\\
            GKEAL\cite{zhuang2023gkeal}&	78.88&	75.62&	72.32&	68.62&	67.23&	64.26&	62.98&	61.89&	60.20&	59.21&	58.67&74.38&+4.86&66.35\\
			\midrule 
			B-FSCL& 77.70& 75.17& 72.72& 67.69& 68.98& 67.63& 67.07& 65.07& 63.36& 63.35& 63.53&81.76&-&68.39 \\
			\bottomrule
			
		\end{tabular}
		\label{tb:fix}  
	}
\end{table*}

\begin{figure*}[!t]
	\footnotesize

	\begin{minipage}[t]{0.32\linewidth}
		\centering
		\includegraphics[width=\linewidth]{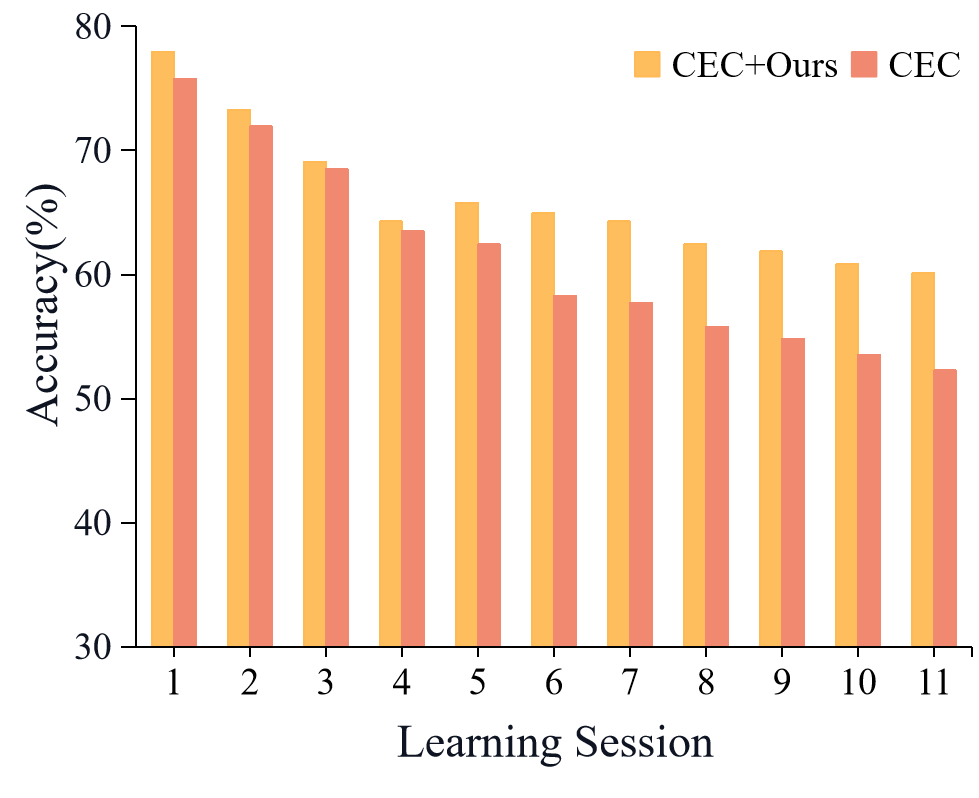}\\
		\vspace{-0.2cm}
		\caption{Performance comparison between CEC+Ours and CEC.}
		\label{fig:integration}\hspace{0.4cm}
	\end{minipage} 
	\begin{minipage}[t]{0.32\linewidth}
		\centering
		\includegraphics[width=\linewidth]{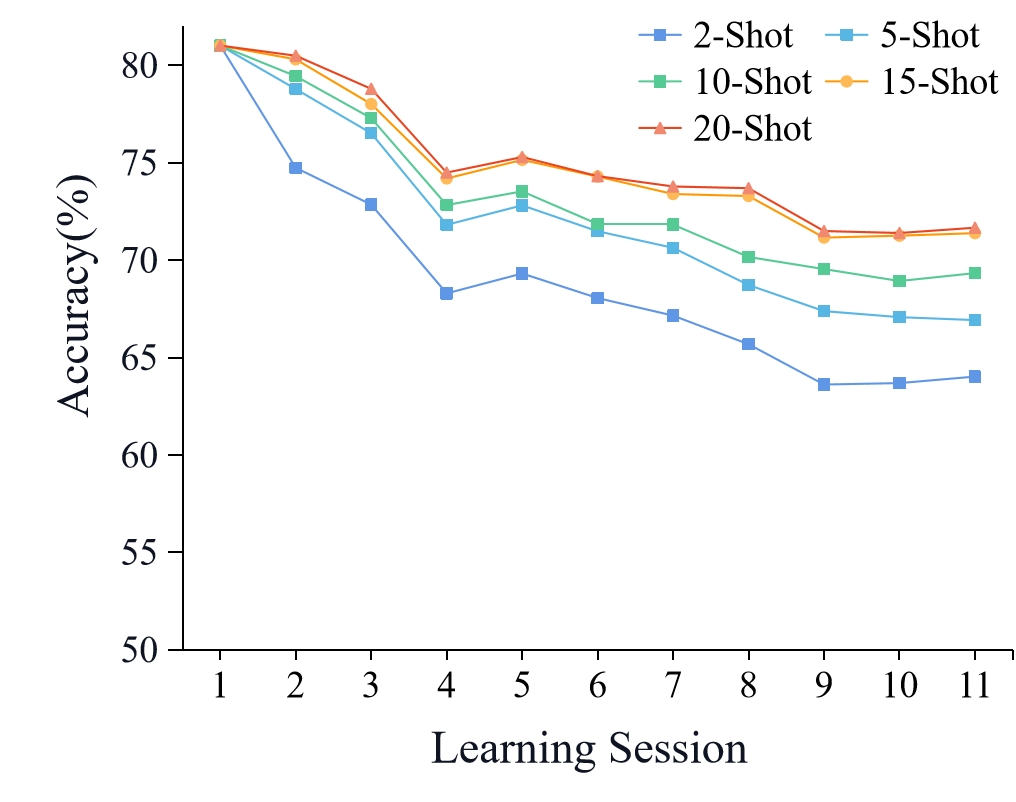}\\
		\vspace{-0.2cm}
		\caption{Performance comparison with the different shot numbers.} \vspace{0cm}
		\label{fig:Kshot}\hspace{0.4cm}
	\end{minipage} 
	\begin{minipage}[t]{0.32\linewidth}
		\centering
		\includegraphics[width=\linewidth]{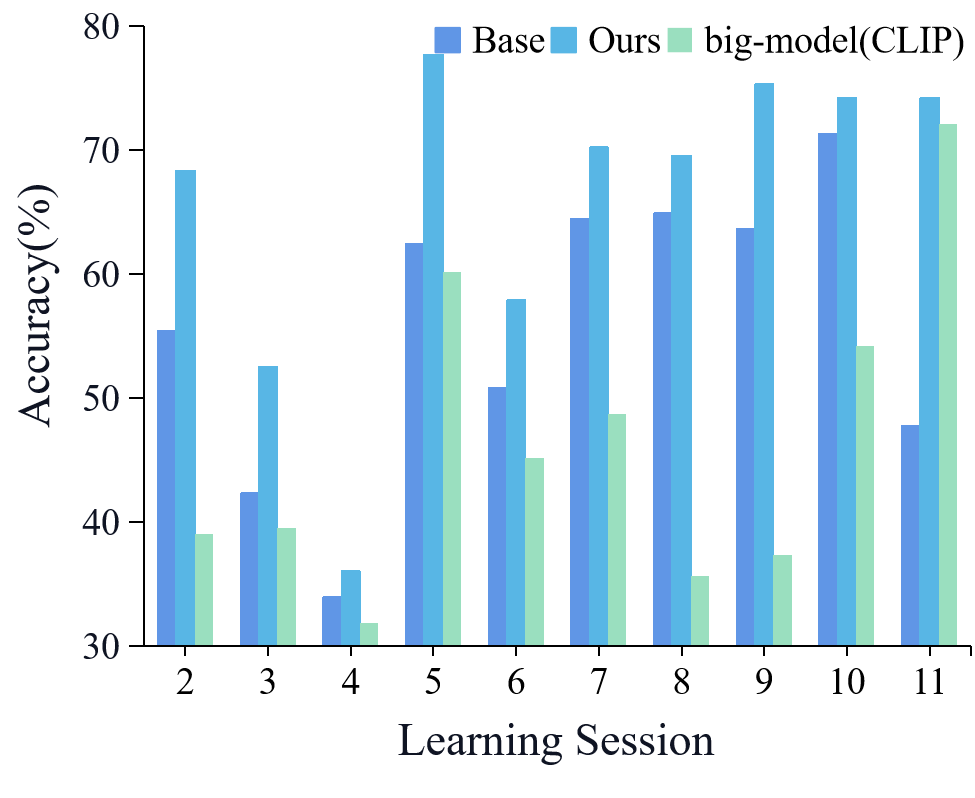}\\
		\vspace{-0.2cm}
		\caption{Performance of new classes in each continual learning session.} \vspace{0cm}
		\label{fig:newclass}
    \end{minipage} 
    \vspace{-0.8cm}
\end{figure*}

\subsection{Ablation study}
All ablation studies are conducted on the CUB200 dataset to assess the efficacy of the proposed modules.

\noindent\textbf{Effectiveness of different components: }
We conduct experiments to assess the effectiveness of the proposed big-model driven embedding transfer (BET) and instance-level adaptive decision (IAD) modules. 
Detailed results are presented in Table~\ref{tb:ablation}.
The performance of the big-model (CLIP) is evaluated at each stage of the continuous learning process. 
The big-model (CLIP) performance remains stable throughout the entire continual learning phase.
Compared to the baseline, the performance of Avg and KR has improved by 2.83\% and 4.22\%, respectively.
These results suggest that our proposed BET module transfers the powerful encoding capability of the big-model to the continual model, thereby enhancing its ability to learn new classes.
Therefore, the efficacy of our proposed big-model driven embedding transfer module is validated.
The main objective of the IAD module is to introduce an adaptive decision mechanism to provide effective inference adjusted to different samples.
This results in significant performance improvements of 8.34\% and 5.34\% on the KR and Avg metrics, respectively.
In conclusion, the integration of both modules results in enhanced performance, establishing a new state-of-the-art performance level. 
This confirms the effectiveness of the proposed B-FSCL framework.

\noindent \textbf{Confusion matrix analysis:}
We further visualize the confusion matrices of our B-FSCL, baseline, and big-model (CLIP) in Fig.~\ref{fig:matrix}.
It is evident that the diagonal elements of the B-FSCL confusion matrix exhibit a darker color compared to the baseline model and big-model (CLIP), clearly indicating that our B-FSCL is more effective at learning and recognizing new classes.
Furthermore, the B-FSCL confusion matrix shows a more concentrated distribution of predictions for the new classes compared to the baseline model and big-model (CLIP), visually highlighting the superior knowledge learning capability of our B-FSCL.

\noindent \textbf{Performance comparison of different methods with similar base performance:}
In Table \ref{tb:fix}, we present a comparison between the B-FSCL framework with three traditional methodologies: GKEAL\cite{zhuang2023gkeal}, ALICE\cite{peng2022few}, and FACT*~\cite{zhou2022forward} where the base class performance is fixed at approximately 78.0\%.
We can observe that the B-FSCL outperforms the other method across all three evaluation metrics, despite keeping the base model performance fixed. 
This implies that the superior performance is not due to the capabilities of the base model, but contributes to the continual model's encoding capacity and the built adaptive decision mechanism.
These results demonstrate that the efficiency of the proposed B-FSCL in acquiring new knowledge while mitigating the challenge of catastrophic forgetting.

\noindent \textbf{Integration of CEC model:}
We perform a comparative experiment by integrating the B-FSCL with CEC~\cite{zhang2021few} presented in Fig.~\ref{fig:integration}.
The results indicate that the performance of CEC is significantly improved by our B-FSCL.
The feature extraction capabilities of the B-FSCL framework are stronger than those of the backbone networks used in CEC.
It proves that the powerful encoding capability of the big-model (CLIP) and the adaptive decision mechanism are leveraged by our B-FSCL, resulting in more effective learning of new knowledge and mitigating catastrophic forgetting.
This confirms the effectiveness of the B-FSCL framework in addressing the problem of few-shot continual learning.
Thereby further confirming the effectiveness of the B-FSCL framework in addressing the problem of few-shot continual learning.

\noindent \textbf{Different shot numbers $K$:}
The accuracy curves with different shot numbers $K$ are presented in Fig.~\ref{fig:Kshot}.
These results demonstrate that employing more samples can significantly improve the performance of the few-shot class continual learning.
This is due to the fact that a larger quantity of samples allows for a more accurate distribution of class attributes, thereby improving the acquisition of new knowledge.
Furthermore, once the data quantity exceeds 15, the rate of performance enhancement diminishes.
It suggests that a satisfactory level of performance can be achieved with an optimal amount of newly annotated data, thereby validating the effectiveness of the B-FSCL to solve the few-shot continual learning task.

\noindent \textbf{Performance of new classes in each continual learning session:}
In order to evaluate the performance of the B-FSCL when learning new classes, we present the accuracy of new classes for the baseline, the B-FSCL, and big-model (CLIP) in each stage of continual learning, as depicted in Fig.~\ref{fig:newclass}.
The experimental results show that our proposed B-FSCL outperform both the baseline and big-model(CLIP) model in the novel class accuracy.
It clearly confirms the superiority of the proposed B-FSCL in effectively learning new knowledge and alleviating catastrophic forgetting.
\vspace{-0.2cm}
\subsection{Conclusion}
In this work, we propose a Big-model driven Few-shot Continual Learning (B-FSCL) framework, which can well employ the strong embedding and reasoning capabilities of the existing big-model to address two key challenges in the FSCL task. 
Under the knowledge traction of the world's big-model, we introduce the big-model driven embedding transfer learning to enhance the convolutional encoding capability of the continual model, which can also mitigate the catastrophic forgetting and over-fitting problems in the continual learning process. 
By facilitating effective communication between the models, we especially design an instance-level adaptive decision module to obtain more accuracy recognition results, which can be used to guide the knowledge distillation for the continual model. 
Extensive experiments have proved the effectiveness of our proposed B-FSCL. 
In future work, we will further employ the advantages of the world's big-model to promote the continual learning ability of artificial neural networks. 
\bibliography{aaai24}
\end{document}